%% file: iclr2025_conference.tex
\documentclass{article} 
\usepackage{iclr2025_conference,times}

\input{math_commands.tex}

\usepackage{hyperref}
\usepackage{url}

\usepackage{amsmath,amssymb,amsfonts}
\usepackage{algorithmic}
\usepackage{graphicx}
\usepackage{textcomp}
\usepackage{xcolor}
\usepackage{multicol}
\usepackage{multirow}
\usepackage{subfig}
\usepackage{booktabs}
\usepackage{enumitem}
\usepackage{caption}

\title{Text-driven Human Motion Generation with Motion Masked Diffusion Model}

\iclrfinalcopy
 
\author{Xingyu Chen\\
University College London\\
\texttt{xingyu.chen.23@ucl.ac.uk} \\
}

\begin{document}

\maketitle

\begin{abstract}
Text-driven human motion generation is a multimodal task that synthesizes human motion sequences conditioned on natural language. It requires the model to satisfy textual descriptions under varying conditional inputs, while generating plausible and realistic human actions with high diversity. Existing diffusion model-based approaches have outstanding performance in the diversity and multimodality of generation. However, compared to autoregressive methods that train motion encoders before inference, diffusion methods lack in fitting the distribution of human motion features which leads to an unsatisfactory FID score. One insight is that the diffusion model lack the ability to learn the motion relations among spatio-temporal semantics through contextual reasoning. To solve this issue, in this paper, we proposed Motion Masked Diffusion Model \textbf{(MMDM)}, a novel human motion masked mechanism for diffusion model to explicitly enhance its ability to learn the spatio-temporal relationships from contextual joints among motion sequences. Besides, considering the complexity of human motion data with dynamic temporal characteristics and spatial structure, we designed two mask modeling strategies: \textbf{time frames mask} and \textbf{body parts mask}. During training, MMDM masks certain tokens in the motion embedding space. Then, the diffusion decoder is designed to learn the whole motion sequence from masked embedding in each sampling step, this allows the model to recover a complete sequence from incomplete representations. Experiments on HumanML3D and KIT-ML dataset demonstrate that our mask strategy is effective by balancing motion quality and text-motion consistency.
\end{abstract}

\section{Introduction}

Generating realistic human motions based on textual descriptions is a complex task due to its multimodal nature. This task bridges the gap between textual semantics and human motion, requires the model not only to accurately understand and translate input text into corresponding motion but also to generate coherent and naturalistic action. However, this task is inherently challenging due to the complexity and variability of human movements. The demand for such technologies is rapidly increasing in various fields, such as graphic animation and human-computer interaction, where generating diverse and contextually relevant motions from textual input can greatly enhance user immersive experience and believable digital content creation.

Existing research has extensively explored various text-driven human motion generation models and algorithms, among them one prevailing method is to construct an motion encoder for natural language inputs~\citep{ahuja2019language2pose, ghosh2021synthesis}.
TEMOS~\citep{petrovich2022temos} trains a transformer variational autoencoder (VAE) architeture to learn the distribution parameters of text-motion latent space from KIT Motion-Language dataset~\citep{plappert2016kit}. 
MotionClip~\citep{tevet2022motionclip} trains a encoder aligned with large pretrained CLIP~\citep{radford2021learning} model. 
After that, T2M-GPT~\citep{zhang2023t2m} introduces Vector Quantised-Variational AutoEncoder (VQ-VAE)~\citep{van2017neural} to learn a discrete motion representation and uses Generative Pretrained Transformer (GPT) for generation with an autoregressive paradigm. 
AttT2M~\citep{zhong2023attt2m} proposes  multi-perspective attention to learn the cross-modal relationship during text-driven motion generation stage. 
MoMask~\citep{guo2024momask} use hierarchical quantization generative model (BERT) to predict the motion sequence based on token generation. 
Typically, these autoregressive architectures learn an motion encoder before generation to capture the semantic representation of the input text, then, a separate decoder or generator model is trained to produce the corresponding motion sequence from these encoded features.

In addition to training motion encoder, there is another approach, Denoising Diffusion Probabilistic Models (DDPM)~\citep{ho2020denoising} for human motion generation. 
MDM~\citep{tevet2023human} have gained prominence due to their ability to generate diverse and high-quality motion sequences in a single-stage process. MDM leverages the many-to-many nature of diffusion processes to generate diverse motion sequences from various forms of conditioning, such as text descriptions or action labels. 
MotionDiffuse~\citep{zhang2022motiondiffuse} leverages diffusion processes to achieve body part independent control with fine-grained texts by multi-level manipulation. 
ReMoDiffuse~\citep{zhang2023remodiffuse} combines retrieval augmented mechanism into diffusion model framework for improving text-motion consistency. 
MLD~\citep{chen2023mld} learns a motion probabilistic mapping in the latent representation space and make the human motion generation more effective on  text-to-motion and action-to-motion tasks.

Comparing to autoregressive architectures such as those built on the GPT~\citep{radford2018improving} or BERT~\citep{kenton2019bert} which follow a two-stage strategy and train a motion decoder before generation. Diffusion models, by contrast, avoid these pitfalls through their single-stage design, where the iterative refinement directly and consistently shapes the motion sequence in line with the text. 
This not only simplifies the training pipeline but also facilitates a more direct and coherent interaction between text and motion representations. 
However, recent research find diffusion architecture often struggle with contextual reasoning~\citep{gao2023masked}, especially in understanding and maintaining relationships among temporal and spatial semantics. 
This limitation arises from their architecture, which, while capable of generating diverse motions, does not inherently prioritize the learning of coherent temporal structures and contextual dependencies necessary for aligning motion with the nuances of language. 

\begin{figure}[t]
\vspace{-10pt}
\centering
    \subfloat[\centering Visualization]
    {{\includegraphics[height=4.42cm]{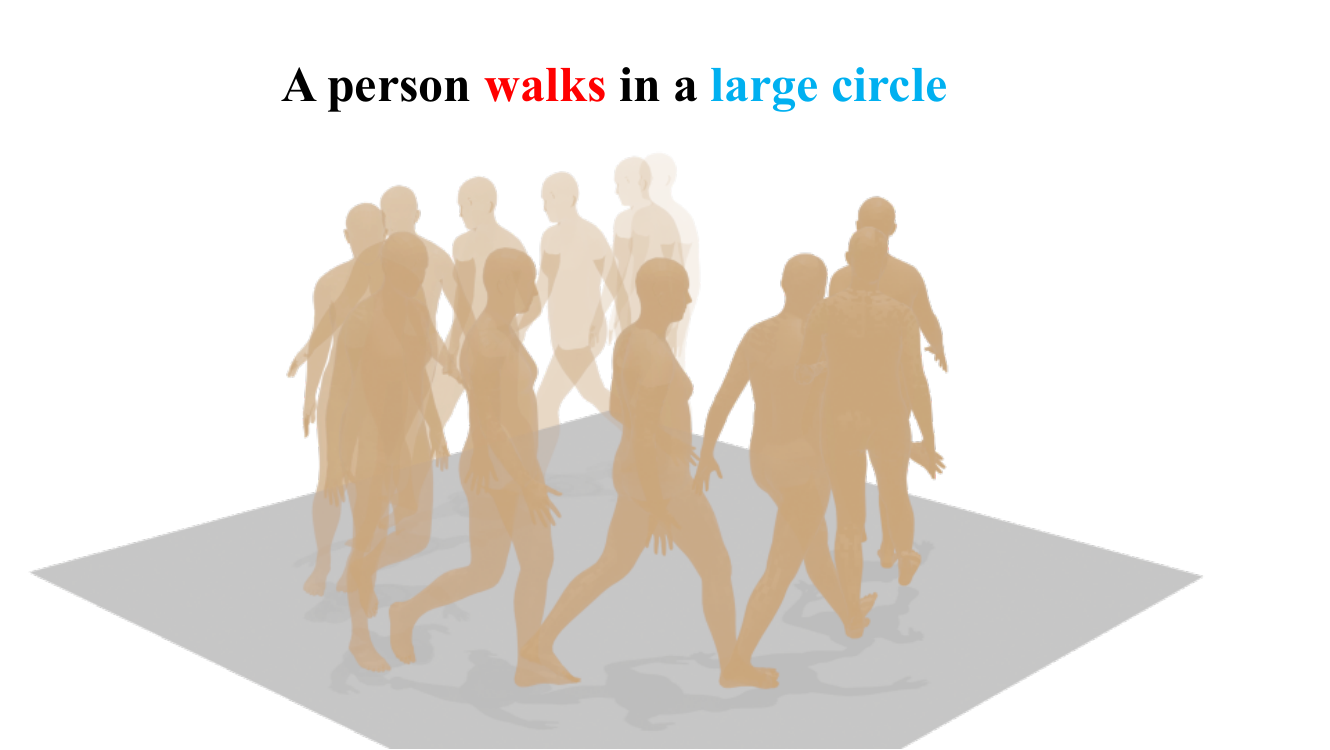} }}%
    \subfloat[\centering Performance]
    {{\includegraphics[height=4.42cm]{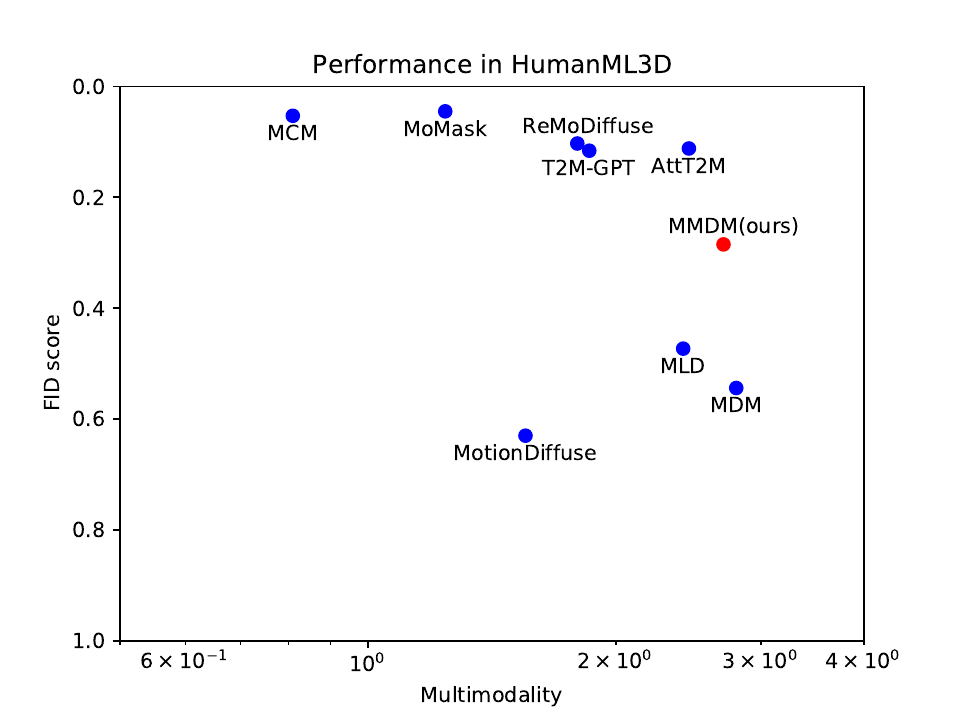} }}%
\caption{
    \textbf{(a) Visualization for text-driven human motion sequence.}  Our method balances the generation quality and diversity of the high-fidelity motion with the semantic consistency of the textual descriptions.
    \textbf{(b) Performance in HumanML3D dataset.} \textit{FID} (lower is better) and \textit{Multimodality} (higher is better) metrics the generation quality and average variance of motion sequences.}
\label{img:01}
\vspace{-10pt}
\end{figure}

To enhance the representation learning capabilities of models, recent advancements in masked strategies~\citep{he2022masked, chang2022maskgit} have shown significant progress. By masking portions of data during training, these models learn to infer robust representations of the missing parts from the visible features, thereby improving their ability to understand and predict contextual relationships. This approach has proven to be particularly effective in diffusion architecture~\citep{gu2022vector, gao2023masked}, where learning to predict masked content fosters a deeper comprehension of the underlying data structure during denoise process. However, directly applying such a strategy into text-to-motion generation is still challenge due to the complexity of human motion data.
Considering the dynamic temporal characteristics and spatial structure of motion, in this paper, we design two motion sequence embedding masked mechanism, time frames mask and body parts mask. During denoising process, MMDM masks certain tokens in the motion embedding space. Then, the decoder is designed to learn the whole motion sequence from masked embedding in each sampling step. MMDM leverages the masked strategy to explicitly improve the model’s capacity to learn spatial and temporal relations. By adopting this strategy, our results indicate that MMDM not only  significantly enhances the coherence and relevance of the motions to the provided textual inputs, but also balances overall quality and diversity of generated motions.

To summarize, the key contributions of our work are as follows:
1) We introduce a embedding space masking strategy into the motion diffusion process, allowing the model to focus on contextual inference of motion generation.
2) We design two Mask mechanism: time frames mask and body parts mask, in order to deal with temporal characteristics and spatial structure of human motion data.
3) Extensive experiment and evaluation on HumanML3D and KIT-ML datasets demonstrating the performance of MMDM, and verify the masked modeling is effective in motion diffusion.

\section{RELATED WORKS}
\label{sec:relaw}

\textbf{Human Motion Synthesis} 
aims to generate realistic 3D human motion sequence. 
The early work focus on unconditional human motion generation~\citep{yan2019convolutional, zhang2020perpetual, zhao2020bayesian}, which random synthesis natural motion sequences from motion capture data without any constraint.
In recent years, research begins to explore the human motion synthesis under various conditions, such as motion prefix~\citep{mao2019learning, liu2022investigating}, action label~\citep{petrovich2021action, guo2020action2motion}, textual description~\citep{guo2022generating, guo2022tm2t, tevet2023human}, image~\citep{rempe2021humor, chen2022learning} or audio signal~\citep{siyao2022bailando, tseng2023edge, gong2023tm2d, zhou2023ude}. Or use the pose sequences as input condition to complete incomplete motion sequences~\citep{harvey2020robust, duan2021single}. 
In this paper, we focus on text-driven motion generation, which is a sub-task of human motion synthesis under textual condition, since the textual descriptors are the convenient and easy to carve out motion details. In the early stage, Text2Action~\citep{ahn2018text2action} employs RNN architecture to train a text to motion mapping from short text. After that, Language2Pose~\citep{ahuja2019language2pose} introduces a concept of Joint Language-to-Pose (JL2P) and applies curriculum learning approach to learn the joint embedding of language and pose. Similarly, MotionCLIP~\citep{tevet2022motionclip} trains this joint embedding space with CLIP encoder. More recent, the current method can be divided into autoregressive architectures~\citep{zhang2023t2m, zhong2023attt2m, guo2024momask} and diffusion architectures~\citep{tevet2023human, zhang2022motiondiffuse, zhang2023remodiffuse, chen2023mld}.

\textbf{Diffusion Generative Models} 
models a stochastic diffusion process by a Markov chain, allowing the model to continuously learn the mapping between each sampling step from the inverse process, leading to denoised generation~\citep{ho2020denoising, dhariwal2021diffusion}. It is also regarded as score-based generative modeling through stochastic differential equations perspective~\citep{song2019generative, song2020improved, song2020score}. Diffusion models combine both elegant physico-mathematical derivations and powerful generative performance, which have attracted plenty of attention from the community. Researchers proposed more efficient sampling strategies DDIM~\citep{song2020denoising}, DPM-solver~\citep{lu2022dpm} to improve diffusion models.
In this paper, we focus on conditional generation for diffusion models. The early work~\citep{dhariwal2021diffusion} applies an extra classifier to guide the gradient during diffusion process. GLIDE~\citep{nichol2021glide} follow this structure and condition on CLIP textual embedding feature. After that, research balance fidelity and diversity, propose Classifier-Free Guidance~\citep{ho2022classifier}, and also align it with CLIP~\citep{ramesh2022hierarchical}. It becomes a dominant generative framework for visual tasks.

\textbf{Generative Masked Modeling} 
is a approach to improve the model ability in learning representations. At early stage, researchers demonstrated the effectiveness of masking methods in natural language processing in representation pretraining stage~\citep{radford2018improving, kenton2019bert} and language generation~\citep{brown2020language}. BERT randomly masked out part of word tokens with a fixed ratio, and use incomplete langauge data to train a bi-directional transformer to predict the masked tokens.
Then computer vision researchers transfer masked modeling approach from NLP area to vision area, and prove its effectiveness~\citep{he2022masked, chang2022maskgit, ji2023masked}. In generative model, masking parts of data are beneficial in generating quality~\citep{zhou2021ibot}, scalability~\citep{he2022masked}, training convergence~\citep{gao2022towards} and contextual reasoning ability~\citep{gao2023masked}. In this paper, we focus on generative masked modeling for human motion generation. 
Momask~\citep{guo2024momask} first introduces BERT architecture and generative masked modeling into human motion synthesis and achieve the state-of-the-art FID score in humanML3D dataset~\citep{guo2022generating}. Inspired by these successes, in this paper, we explore generative masked modeling approach for human motion diffusion models.

\section{METHODOLOGY}
\label{sec:methd}

\begin{figure*}[t]
    \centering
    \includegraphics[width = 1\textwidth]{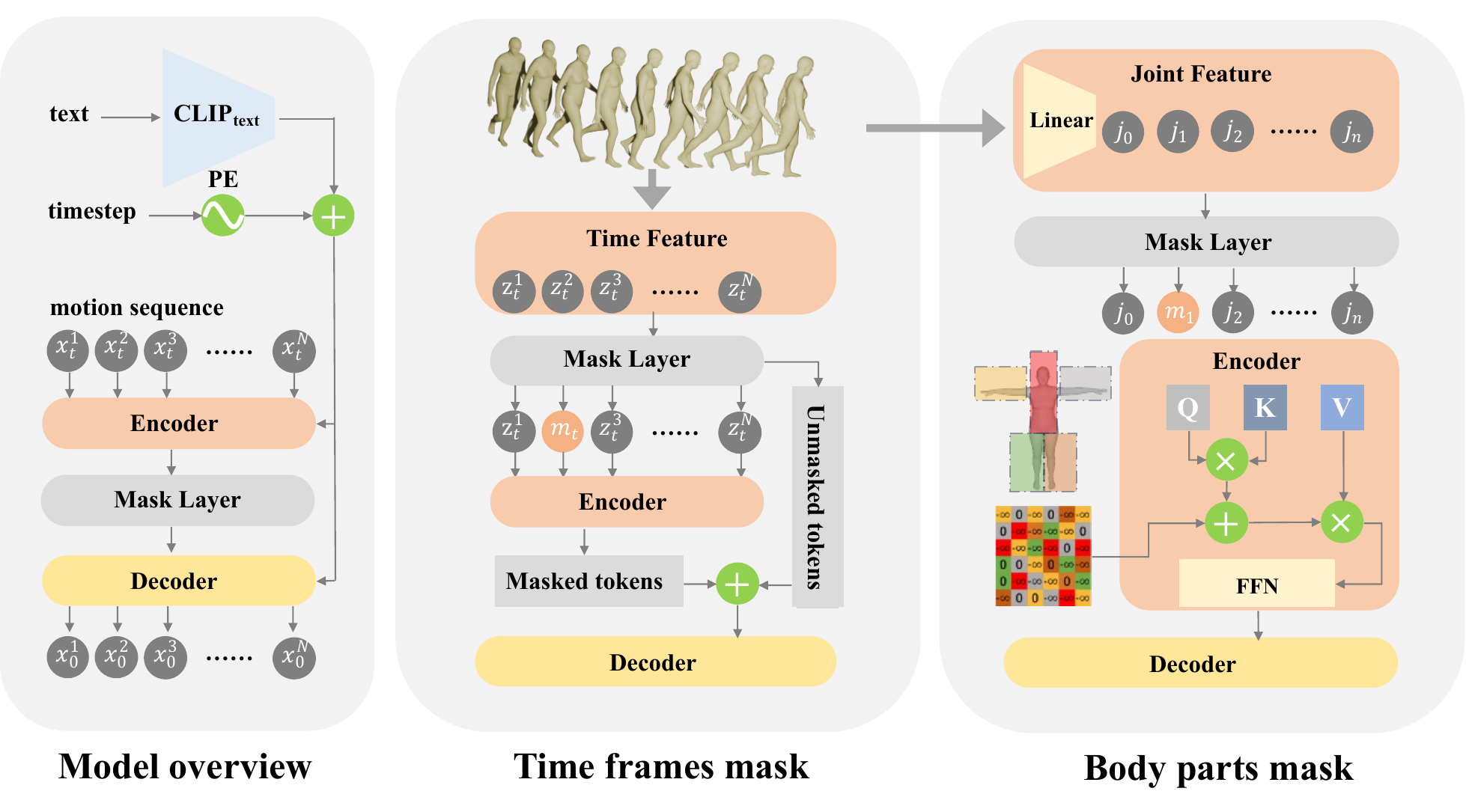}
    \caption{\textbf{Overview and network}. We propose motion mask diffusion model (Left), including time frames mask (Middle) and body parts mask (Right) for expressive spatio-temporal features in the motion embedding. 
    Given a natural language condition, a CLIP~\citep{radford2021learning} transfer it into textual embedding and projected together with positional embedding of timestep $t$ for classifier-free learning~\citep{ho2022classifier}. In each sampling step, our model follow MDM~\citep{tevet2023human} directly predicts the final clean motion sequence ${x}^{1:N}$ instead of noise, then repeats from $x_t$ to $x_0$.}
    \label{img:overview}
\end{figure*}

In this paper, we demonstrate a Motion Masked Diffusion Model (MMDM). An overview framework of our method is presented in Figure \ref{img:overview}.
We introduces a masked embedding modeling scheme into the diffusion process to enhance the contextual reasoning ability for human motion generation.

\subsection{human motion diffusion model}

Motion Masked Diffusion Model (MMDM) proposed by \citet{tevet2023human},  enables the diffusion model to learn text to motion representations for generation. Given an arbitrary text condition $c$, the purpose of MDM is to synthesize a human motion $X = [x^1, x^2, x^3, ..., x^N]$ with length $N$, where $N$ is the number of time frames. The output human motion $x^{1:N}=\{x^i\}_{i=1}^{N}$ is a sequence of human poses represented by either joint positions or rotations with $x^i \in \mathbb{R}^{J \times D}$, where $J$ is the number of joints and $D$ is the dimension of the joint representations.

\textbf{Diffusion Process}.
The diffusion probabilistic models is modeled by a Markov chain, and involves a forward noise adding process and a reverse denoising process. For sampling sequence $\{x^{1:N}_{t}\}_{t=0}^T$,  where $x^{1:N}_{0}$ is drawn from the data distribution and
\begin{equation}
{
q(x^{1:N}_{t} | x^{1:N}_{t-1}) = \mathcal{N}(\sqrt{\alpha_{t}}x^{1:N}_{t-1},(1-\alpha_{t})I),
}
\end{equation}
where $\alpha_{t} \in (0,1)$ are constant hyper-parameters.
In human motion data, conditioned motion synthesis models the distribution $p(x_{0}|c)$  as the reversed diffusion process of gradually cleaning $x_{T}$. 

\begin{equation}
{
\mathcal{L}_\text{simple} = E_{x_0 \sim q(x_0|c), t \sim [1,T]}[\| x_0 - \hat{x}_{0}\|_2^2]
}
\end{equation}

Instead of predicting $\epsilon_t$ then add it into sampling data as formulated by \citet{ho2020denoising}, we follow \citet{ramesh2022hierarchical} and predict the human motion sequences itself~\citep{tevet2023human}.
This allows more geometric constraints to be added directly to the optimization function for controlling more realistic human motion generation.

\textbf{Optimization function}.
We follow the standard geometric regularization for human motion domain from~\cite{petrovich2021action, shi2020motionet}.
1) Position Loss: This loss ensures that the predicted joint positions are consistent with the true joint positions. 2) Foot Contact Loss: To prevent foot sliding and maintain realistic foot-ground interactions, this term penalizes deviations in foot position during contact with the ground. 3) Velocity Loss: This loss encourages smooth and consistent motion by regulating the velocity of the joints.

\begin{equation}
{
\mathcal{L}_\text{pos} = \frac{1}{N} \sum_{i =1}^{N} \| FK(x^i_0) - FK(\hat{x}^i_0)\|_{2}^{2}, %
}
\end{equation}
where \( FK(\cdot) \) is the forward kinematics function that converts joint rotations into positions.

\begin{equation}
{
\mathcal{L}_\text{foot} = \frac{1}{N-1} \sum_{i =1}^{N-1} \| (FK(\hat{x}^{i+1}_0) - FK(\hat{x}^i_0)) \cdot f_i\|_{2}^{2},
}
\end{equation}
where \( f_i \) is a binary indicator of foot contact at frame \( i \).

\begin{equation}
{
\mathcal{L}_\text{vel} = \frac{1}{N-1} \sum_{i =1}^{N-1} \| (x^{i+1}_0 - x^i_0) - (\hat{x}^{i+1}_0 - \hat{x}^i_0)\|_{2}^{2}
}
\end{equation}

The final training loss combines these geometric losses with the standard diffusion loss:
\begin{equation}
{
\mathcal{L} = \mathcal{L}_\text{simple} + \lambda_\text{pos}\mathcal{L}_\text{pos} + \lambda_\text{vel}\mathcal{L}_\text{vel} + \lambda_\text{foot}\mathcal{L}_\text{foot}.
}
\end{equation}
where \( \lambda_{pos} \), \( \lambda_{foot} \), and \( \lambda_{vel} \) are weighting factors that balance the contributions of the geometric losses.

\subsection{Motion embedding Masked Diffusion}

To enhance the contextual reasoning ability of diffusion models, we introduce masking strategy within the embedding space during the diffusion process, termed MMDM. The core idea of Motion Masked Diffusion Model is randomly masking a subset of the input in order to force the model reasoning about missing parts based on incomplete data, which can lead to a more robust and contextually aware model.
Specifically, at each time step, a certain proportion of the embedding variables corresponding to the motion sequence are masked, effectively hiding parts of the motion from the model. The model is then tasked with reconstructing the entire sequence, including the masked components, from the remaining visible parts. This forces the model to learn to infer the missing context, thereby enhancing its ability to understand and predict the complex relationships between spatial and temporal semantics in human motion.

Given a motion embedding \( z_{1:N} \) of length \( N \), where each frame \( z_i \) represents the features of the time frames or pose joints at time step \( i \). At the start of the diffusion process, a mask \( M \) is initialized by randomly selecting with mask ratio, and a learnable embedding \( q \) is initialized by mask token adding positional encoding. The mask can be designed to either randomly select time frames to mask at each step or to focus on specific joints known to be critical for body parts. This depends on the mask type, it will be introduced in the next section. Mask token can ensure the decoder always receives same size of motion embedding during training and inference. At each diffusion step \( t \), noise is added to the motion sequence based on the mask \( M \). The noisy sequence \( \tilde{x}_{1:N}^t \) at step \( t \) is given by:
\[
\tilde{x}_{1:N}^t = M \odot q + (1 - M) \odot x_{1:N}^{t-1}
\]
where \( \odot \) denotes element-wise multiplication. The mask \( M \) ensures that mask token is only applied to the selected parts of the motion embedding, allowing the model to focus on both denoising and contenxtual reasoning.

Traditional mask modeling apply mask token to the parts of input data, however, the human motion sequence is high dimensions data with three dimensions in space and one dimension in time frames. The key innovation in MMDM lies in the masked strategies for human motion data. Consider its complexity in spatial structure and dynamic temporal characteristics, we design two mask modeling: time frames mask and body parts mask.

\subsubsection{Time frames mask}

For time frames mask, we introduces the asymmetric diffusion transformer\citep{gao2023masked} which including an encoder and a decoder for joint training of mask embedding modeling and diffusion process. During training, the encoder processes masked embedding and predict the full tokens, and the masked tokens from encoder prediction will be added together with the unmasked portion to be fed into the decoder for diffusion training shown as Figure\ref{img:overview} middle part.

\textbf{Time frames encoder}.
Firstly, the motion embedding takes masking operation on the time dimension, if a time frame is selected, it will be substituted by a learnable mask token. Secondly, to enhance the relative position information during learning, positional encoding is necessary for all motion embedding. In our model, the encoder adds the conventional learnable global position embedding into the noisy embedding latent input. Thirdly, a transformer block is designed for computing the attention score of self-attention:

\begin{equation}
\text{Attention}(Q, K, V) = \text{softmax} \left( \frac{Q K^\top}{\sqrt{d_k}} + B_r \right) V
\end{equation}

where $Q$, $K$, and $V$, respectively denote the query, key, and value in the self-attention module, $d_k$ is the dimension of the key, and $B_r$ is the relative positional bias~\citep{liu2021swin}. The relative positional bias is helpful to capture the token relationship, facilitating the prediction on masked tokens. 

\textbf{Time frames decoder}.
Similar to encoder, the decoder also adds learnable position embedding into its input motion tokens to enhance positional information. But the difference between them is that the encoder is concerned only with the unmasked portion, while the decoder needs to be concerned with all tokens, so this structure is asymmetric diffusion transformer architecture\citep{gao2023masked}.

\subsubsection{Body parts mask}
For body parts mask, we introduces Body-Part attention-based Spatio-Temporal(BPST) encoder ~\citep{zhong2023attt2m} for learning the body parts features. Then we mask out a portion of the body parts features for diffusion training shown as Figure\ref{img:overview} right part.

\textbf{Body parts encoder}.  
Firstly, we follow the BPST encoder~\citep{zhong2023attt2m} divide the human body skeleton with n joints into five body parts: \textit{Torso, Left Arm, Right Arm, Left Leg, Right Leg}, each containing its own set of joints. 
Secondly, a linear layer is applied for mapping all tokens into the same dimension before computing self-attention. 
Thirdly, we construct a adjacency matrix $M$ from the joint information. For example, if a joint $i$ and joint $j$ belong to the same body parts $m_{i,j} = 0$, otherwise $m_{i,j} = - \infty$.
Then, we concatenate the mapping results of different tokens after obtaining the feature from parts to compute the body-part attention:

\begin{equation}
\text{Attention}(Q, K, V) = \text{softmax} \left( \frac{Q K^\top + M}{\sqrt{d_k}}  \right) V
\end{equation}

where $Q$, $K$, and $V$, respectively denote the query, key, and value in the self-attention module, $d_k$ is the dimension of the key, and $M$ is body parts adjacency matrix~\citep{zhong2023attt2m}. After obtaining the final spatial feature of body parts, we takes mask operation for randomly select portion of body parts feature and substitute with learnable mask tokens.

\textbf{Body parts decoder}. Similar to time frames decoder, we also adds position encoding to learn the position information. However, unlike the masking operation on the previous time frames mask, the masking operation on the body part features is executed after encoder. So the body parts decoder needs to complete the mask motion data and take diffusion learning at the same time.

\section{EXPERIMENTS}
\label{sec:exper}

\begin{table*}[t]
\begin{center}
\resizebox{\textwidth}{!}{
    \begin{tabular}
    {p{2.15cm}p{1.70cm}p{1.70cm}p{1.70cm}p{1.70cm}p{1.70cm}p{1.70cm}p{1.70cm}}
    \hline
    \multirow{2}{*}{Methods}&\multicolumn{3}{c}{R Precision$\uparrow$}&\multirow{2}{*}{FID.$\downarrow$}&\multirow{2}{*}{MM-D.$\downarrow$}& \multirow{2}{*}{Div.$\rightarrow$}& \multirow{2}{*}{MM.$\uparrow$}\\ \cline{2-4}
    & Top-1& Top-2& Top-3&&&&\\
    \hline
    
    Real& 0.511 \scriptsize $\pm$ .003& 0.703  \scriptsize $\pm$ .003& 0.797  \scriptsize $\pm$ .002& 0.002  \scriptsize $\pm$ .000&2.974  \scriptsize $\pm$ .008 &9.503 \scriptsize $\pm$ .065 &-\\
    
    \hline
    T2M & 0.455 \tiny $\pm$ .003& 0.636  \tiny $\pm$ .003& 0.736 \tiny $\pm$ .002& 1.087 \tiny $\pm$ .021&3.347  \tiny $\pm$ .008 &9.175 \tiny $\pm$ .083 &2.219 \tiny $\pm$ .074\\
    
    MDM & 0.320 \scriptsize $\pm$ .005& 0.498  \scriptsize $\pm$ .004& 0.611  \scriptsize $\pm$ .007& 0.544  \scriptsize $\pm$ .044&5.566 \scriptsize $\pm$ .027 &\textbf{9.559} \scriptsize $\pm$ .086 &\textbf{2.799} \scriptsize $\pm$ .072\\
    
    MotionDiffuse & \textbf{0.491} \scriptsize $\pm$ .001& \textbf{0.681}  \scriptsize $\pm$ .001& \textbf{0.782}  \scriptsize $\pm$ .001& 0.630  \scriptsize $\pm$ .001&\textbf{3.113}  \scriptsize $\pm$ .001 &\underline{9.410} \scriptsize $\pm$ .049 &1.553 \scriptsize $\pm$ .042\\
    
    MLD & \underline{0.481} \scriptsize $\pm$ .003& 0.673 \scriptsize $\pm$ .003& 0.772  \scriptsize $\pm$ .002& 0.473  \scriptsize $\pm$ .013&3.196  \scriptsize $\pm$ .010 &9.724 \scriptsize $\pm$ .082 &2.413 \scriptsize $\pm$ .079\\ 
    
    T2M-GPT & \textbf{0.491} \scriptsize $\pm$ .003& \underline{0.680}  \scriptsize $\pm$ .003& \underline{0.775}  \scriptsize $\pm$ .002& \textbf{0.116}  \scriptsize $\pm$ .004&\underline{3.118}  \scriptsize $\pm$ .011 &9.761 \scriptsize $\pm$ .081 &1.856 \scriptsize $\pm$ .011\\
    
    \hline
    
    MMDM-t & 0.464 \scriptsize $\pm$ .006& 0.654 \scriptsize $\pm$ .007& 0.754  \scriptsize $\pm$ .005& 0.319  \scriptsize $\pm$ .026&3.288  \scriptsize $\pm$ .023 &9.299 \scriptsize $\pm$ .064 &\underline{2.741} \scriptsize $\pm$ .112\\ 

    MMDM-b & 0.435 \scriptsize $\pm$ .006& {0.627}  \scriptsize $\pm$ .006& {0.733}  \scriptsize $\pm$ .007& \underline{0.285} \scriptsize $\pm$ .032&{3.363} \scriptsize $\pm$ .029 &{9.398} \scriptsize $\pm$ .088 &{2.701} \scriptsize $\pm$ .083\\
    \hline
    \end{tabular}}
\end{center}
\caption{\textbf{Quantitative evaluation on the testset of HumanML3D.}
We report the metrics following T2M~\citep{guo2022generating} and repeat 20 times to get the average results with 95\% confidence interval. The $\downarrow, \uparrow, and \rightarrow$ denote the lower, higher, and closer to Real are better, respectively. The best results are marked in bold and the second best is underlined. Our method achieves significant improvement on almost all metrics.}
\label{table:01}
\end{table*}

\begin{table*}[t]

\begin{center}
\resizebox{\textwidth}{!}{
    \begin{tabular}
    {p{2.15cm}p{1.70cm}p{1.70cm}p{1.70cm}p{1.70cm}p{1.70cm}p{1.70cm}p{1.70cm}}
    \hline
    \multirow{2}{*}{Methods}&\multicolumn{3}{c}{R Precision$\uparrow$}& \multirow{2}{*}{FID.$\downarrow$}&\multirow{2}{*}{MM-D.$\downarrow$}& \multirow{2}{*}{Div.$\rightarrow$}& \multirow{2}{*}{MM.$\uparrow$}\\ \cline{2-4}
    &Top-1&Top-2&Top-3&&&&\\
    \hline
    
     Real& 0.424 \tiny $\pm$ .005& 0.649  \tiny $\pm$ .006& 0.779  \tiny $\pm$ .006& 0.031  \tiny $\pm$ .004&2.788  \tiny $\pm$ .012 &11.08 \tiny $\pm$ .097 &-\\
     \hline
    
    T2M & 0.361 \tiny $\pm$ .006& 0.559  \tiny $\pm$ .007& 0.681 \tiny $\pm$ .007& 3.022 \tiny $\pm$ .107&3.488  \tiny $\pm$ .028 &10.72 \tiny $\pm$ .145 &2.052 \tiny $\pm$ .107\\
    
    MDM & 0.164 \tiny $\pm$ .004& 0.291  \tiny $\pm$ .004& 0.396  \tiny $\pm$ .004& {0.497}  \tiny $\pm$ .021&9.191 \tiny $\pm$ .022 &10.85 \tiny $\pm$ .109 &1.907 \tiny $\pm$ .214\\
    
    MotionDiffuse & \textbf{0.417} \tiny $\pm$ .004& \underline{0.621}  \tiny $\pm$ .004& \underline{0.739}  \tiny $\pm$ .004& 1.954 \tiny $\pm$ .064&\underline{2.958} \tiny $\pm$ .005 &\textbf{11.10} \tiny $\pm$ .143 & 0.730\tiny \ $\pm$ .013\\
    
    MLD & 0.390 \tiny $\pm$ .008& 0.609 \tiny $\pm$ .008& 0.734  \tiny $\pm$ .007& \underline{0.404}  \tiny $\pm$ .027&3.204  \tiny $\pm$ .027&10.80 \tiny $\pm$ .117 &\underline{2.192} \tiny $\pm$ .071\\
    
    T2M-GPT& 0.402 \tiny $\pm$ .006& 0.619  \tiny $\pm$ .005& 0.737  \tiny $\pm$ .006& 0.717 \tiny $\pm$ .041&3.053  \tiny $\pm$ .026 &\underline{10.86} \tiny $\pm$ .094 &1.912 \tiny $\pm$ .036\\
    
    \hline
    
    MMDM-t & \underline{0.432} \scriptsize $\pm$ .006& \textbf{0.643} \scriptsize $\pm$ .007& \textbf{0.760}  \scriptsize $\pm$ .006& \textbf{0.237}  \scriptsize $\pm$ .013&\textbf{2.938}  \scriptsize $\pm$ .025 &10.84\scriptsize $\pm$ .125 &1.457 \scriptsize $\pm$ .129\\ 

    MMDM-b & 0.386 \scriptsize $\pm$ .007& 0.603 \scriptsize $\pm$ .006& 0.729  \scriptsize $\pm$ .006& 0.408  \scriptsize $\pm$ .022&3.215  \scriptsize $\pm$ .026 &10.53 \scriptsize $\pm$ .100 &\textbf{2.261} \scriptsize $\pm$ .144\\ 
    
    \hline
    \end{tabular}}
\end{center}
\caption{\textbf{Quantitative evaluation on the testset of KIT-ML.} The experimental settings are the same as Table~\ref{table:01}.We report the metrics following T2M~\cite{guo2022generating} and repeat 20 times to get the average results with 95\% confidence interval. The best results are marked in bold and the second best is underlined.}
\label{table:02}
\end{table*}

We implement MMDM using the PyTorch framework. The time-frames mask transformers are initialized with 2 layers encoder and 6 layers decoder with a hidden dimension of 512. The body-parts mask transformer decoder is initialized with 6 layers and a hidden dimension of 640. 
We train the model using the Adam optimizer with a learning rate of \(10^{-4}\) and a batch size of 64. The cosine noise schedule is applied during the diffusion process, with 1,000 noising steps. 
For the masking mechanism, the mask token is dynamically updated based on the gradients of the loss function during training, allowing the model to learn the most important frames or joints for accurate generation.

\subsection{Set up}

\textbf{Dataset.} We conduct our experiments on two widely used human motion datasets:
\begin{itemize}
    \item HumanML3D \citep{guo2022generating}: This dataset contains 14,616 motion sequences annotated with 44,970 textual descriptions. It includes a wide variety of human actions and motions, providing a comprehensive benchmark for text-to-motion generation tasks.
    \item KIT Motion-Language \citep{plappert2016kit}: The KIT dataset consists of 3,911 motion sequences paired with textual descriptions. Although smaller in size compared to HumanML3D, it is commonly used in text-to-motion research, making it an important benchmark for evaluating model performance.
\end{itemize}

\textbf{Evaluate metrics.} To quantitatively evaluate the performance of our model, we use the following metrics:
{Fréchet Inception Distance (FID)}: This metric measures the similarity between the distribution of generated motions and the ground truth motions. Lower FID scores indicate better performance.
{R-Precision}: This metric evaluates the relevance of generated motions to the input textual descriptions by computing the top-k accuracy of the retrieval results. Higher R-Precision scores indicate better alignment with the input text.
{Diversity}: This metric assesses the variability in the generated motion sequences, ensuring that the model does not collapse to generating a limited set of motions.
{Multimodality}: This metric measures the average variance of generated motions given a single text prompt, reflecting the model's ability to generate diverse outputs from the same input.

\textbf{Quantitative evaluation.} 
Table \ref{table:01} and Table \ref{table:02} demonstrate MMDM performance in the text-driven human motion diffsuion task on the HumanML3D and KIT datasets respectively. We conduct 20 evaluations, with 1000 samples in each, and report their average and a 95\% confidence interval.

\subsection{Results and Analysis}

\subsection{Ablation Study on Masking Mechanism}

\begin{table}[h]
  \begin{center}
  \small
  \renewcommand{\arraystretch}{1.1} 
  \setlength{\tabcolsep}{2mm} 
    \begin{tabular}{c|ccccccc}
        \toprule
        Dataset & Mask Type & Ratio &  FID$\downarrow$ & Top-3 R Precision$\uparrow$ & MM-D.$\downarrow$ & Div.$\rightarrow$ & MM.$\uparrow$ \\	\midrule

        \multirow{8}{*}{HumanML3D} & \multirow{4}{*}{MMDM-t} 
        &0.1 & 0.286 & \textbf{0.743} & 3.383 & \textbf{9.361} & \textbf{2.795}  \\
         ~ & ~ &0.2 & \textbf{0.276} & 0.742 & \textbf{3.355} & 9.285 & 2.741   \\
         ~ & ~ &0.3 & 0.302 & 0.734 & 3.426 & 9.288 & 2.674 \\
         ~ & ~ &0.4 & 0.349 & 0.733 & 3.422 & 9.144 & 2.661 \\
        \cline{2-8}
        ~ & \multirow{4}{*}{MMDM-b} 
        &0.1 & \textbf{0.252} & \textbf{0.744} & \textbf{3.338} & \textbf{9.442} & 2.701  \\
        ~ & ~ &0.2 & 0.614 & 0.712 & 3.588 & 8.765 & \textbf{2.832}  \\
        ~ & ~ &0.3 & 1.539 & 0.689 & 3.723& 8.409 & 2.803 \\
        ~ & ~ &0.4 & 1.542 & 0.677 & 3.810 & 8.424 & 2.829 \\
        
        \midrule 

        \multirow{8}{*}{KIT}  & \multirow{4}{*}{MMDM-t} 
        &0.1 & \textbf{0.234} & 0.767 & 2.937 & 10.77 & 1.535  \\
         ~ & ~ &0.2 & 0.278 & \textbf{0.772} & \textbf{2.925} & \textbf{10.84} & 1.500   \\
         ~ & ~ &0.3 & 0.328 & 0.770 & 3.004 & 10.74 & \textbf{1.567} \\
         ~ & ~ &0.4 & 0.366 & 0.749 & 3.038 & 10.72 & 1.550 \\
        \cline{2-8}
        ~ & \multirow{4}{*}{MMDM-b} 
        &0.1 & 0.923 & 0.664  & 3.755 & 10.21 & \textbf{2.695} \\
        ~ & ~ &0.2 & \textbf{0.449} & \textbf{0.735} & \textbf{3.196} & 10.49 & 2.261 \\
        ~ & ~ &0.3 & 0.481 & 0.714 & 3.312 & 10.40 & 2.440 \\
        ~ & ~ &0.4 & 0.697 & 0.693 & 3.550 & \textbf{10.57} & 2.399 \\
        
        \bottomrule
    \end{tabular}
    \end{center}
    \caption{\textbf{Effect of different masking ratios.} We repeat 5 times to get the average results with 95\% confidence interval under different mask ratio. The $\downarrow, \uparrow, and \rightarrow$ denote the lower, higher, and closer to Real are better, respectively.}
    \label{tab:maskrato}
    \vspace{-4pt}
\end{table}

To analyze the impact of the masking mechanism, we conduct an ablation study where we compare the performance of MMDM with different mask ratio. As the mask rate increases, the model focuses more on the mask portion rather than diffusion generation. In Table \ref{tab:maskrato}, Diffusion masking on time frames or on body part features, the optimal masking ratio are both in the interval 0.1-0.2.

\subsection{Ablation Study on Model Architecture}

\begin{table}[h]
  \begin{center}
  \small
  \setlength{\tabcolsep}{3mm} 
    \begin{tabular}{ccccccc}
        \toprule
        Arch &  FID$\downarrow$ & Top-3 R Precision$\uparrow$ & MM-D.$\downarrow$ & Div.$\rightarrow$ & MM.$\uparrow$ \\	\midrule
        04 Encoder+2 Decoder & 0.461 & 0.721 & 3.469 & 9.136 & 2.684   \\
        \textbf{06 Encoder+2 Decoder} & \textbf{0.232} & \textbf{0.746} & \textbf{3.333} & 9.335 & 2.596    \\
        08 Encoder+4 Decoder & 0.296 & 0.742 & 3.405 & 9.090 & \textbf{2.694}   \\
        12 Encoder+4 Decoder & 0.369 & 0.731 & 3.399 & \textbf{9.442} & 2.638  \\
        \bottomrule
    \end{tabular}
    \end{center}
    \caption{\textbf{Effect of different architecture.} We repeat 5 times to get the average results under 95\% confidence interval with mask ratio 0.2 in HumanML3D dataset. The $\downarrow, \uparrow, and \rightarrow$ denote the lower, higher, and closer to Real are better, respectively.}
    \label{tab:maskarch}
    \vspace{-4pt}
\end{table}

To analyze the impact of the masking mechanism, we also conduct an ablation study where we compare the performance under different number of layers. As shown in Table \ref{tab:maskarch}, We tested four different sets of model capacities for time frames mask. From the FID score, the model size of a 6-layer encoder with a 2-layer decoder is optimal for the HumanML3D dataset. For a fair comparison, we followed this layer structure for our experiments on the body parts mask and KIT dataset.

\section{Conclusion}
\label{sec:concls}

In this paper, we introduced Motion Masked Diffusion model (MMDM), a novel approach that integrates the generative masking strategy into the Human Motion Diffusion Model. Our proposed model addresses key challenges in human motion diffusion model to improve its FID-score performance, while maintaining diversity and realism in generated motions.

\bibliography{iclr2025_conference}
\bibliographystyle{iclr2025_conference}


\end{document}

%% file: math_commands.tex
\usepackage{amsmath,amsfonts,bm}

\def\eqref#1{equation~\ref{#1}}

\def\1{\bm{1}}

\DeclareMathAlphabet{\mathsfit}{\encodingdefault}{\sfdefault}{m}{sl}
\SetMathAlphabet{\mathsfit}{bold}{\encodingdefault}{\sfdefault}{bx}{n}

